\documentclass{article}
\PassOptionsToPackage{table}{xcolor}
\usepackage[utf8]{inputenc}
\usepackage{algorithmic}

\usepackage{microtype}
\usepackage{graphicx}
\usepackage{subcaption}
\usepackage{booktabs} 

\usepackage{makecell}

\usepackage{hyperref}
\usepackage{algorithm}
\makeatletter
\expandafter\let\csname algorithmic\endcsname\relax
\expandafter\let\csname endalgorithmic\endcsname\relax
\makeatother
\usepackage{algpseudocode}

\newcommand{\piLLM}{\pi_{\theta_{\mathrm{LLM}}}}
\newcommand{\QD}{\mathrm{QD}}

\usepackage[accepted]{icml2026}

\usepackage{amsmath}
\usepackage{amssymb}
\usepackage{mathtools}
\usepackage{amsthm}
\usepackage{xcolor}

\usepackage[capitalize,noabbrev]{cleveref}

\theoremstyle{plain}

\theoremstyle{definition}

\theoremstyle{remark}

\usepackage[textsize=tiny]{todonotes}
\usepackage{xcolor}

\icmltitlerunning{GAE: Graph-Augmented Evolution for Scientific Discovery
via Reinforcement Optimization}

\begin{document}

\twocolumn[
  \icmltitle{\textbf{GAE}: Graph-Augmented Evolution for Scientific Discovery \\
       via Reinforcement Optimization}



  \icmlsetsymbol{equal}{*}

  \begin{icmlauthorlist}
    \icmlauthor{Xuanzhou Chen}{sais,gat}
    \icmlauthor{Taoli Cheng}{sais}

  \end{icmlauthorlist}

  \icmlaffiliation{sais}{Shanghai Academy of AI for Science, Shanghai, China}
  \icmlaffiliation{gat}{School of Electrical and Computer Engineering, Georgia Institute of Technology, Georgia, Unites States}

  \icmlcorrespondingauthor{Taoli Cheng}{chengtaoli.1990@gmail.com}

  \icmlkeywords{Machine Learning, ICML}

  \vskip 0.3in
]



\printAffiliationsAndNotice{}  

\begin{abstract}

Evolutionary program search guided by Large Language Models (LLMs) has emerged as a powerful paradigm for automated scientific discovery. However, current approaches are fundamentally constrained by three bottlenecks: structurally blind parent selection, sparse whole-program evaluation rewards, and static mutation operators that fail to adapt during search. We present \textbf{GAE} (\textbf{G}raph-\textbf{A}ugmented \textbf{E}volution), a framework that resolves these limitations through a tightly coupled, three-pillar architecture. First, a \textbf{relational graph neural network (GNN)} parses programs into typed computation graphs, producing structure-aware embeddings. Second, an \textbf{RL-optimized meta-controller} leverages these embeddings to replace blind evolutionary sampling with a directed policy, dynamically selecting optimal parents and mutation directions based on reward history. Third, an \textbf{online GRPO fine-tuning loop} continuously updates the LLM mutation operator at test-time using group-normalized evaluation rewards, directly aligning the model's generation distribution with high-fitness structural edits. We evaluate GAE on a challenging scientific discovery task: symbolic regression for complex nonlinear oscillator systems. By transforming stochastic search into a directed, self-improving trajectory, GAE efficiently discovers closed-form physical equations, consistently matching or outperforming static LLM-driven baselines and achieving state-of-the-art out-of-distribution performance.
  
\end{abstract}

\section{Introduction}
The discovery of new algorithms and programs is a central frontier in AI-assisted science. 
Recent work follows a simple but powerful recipe: combine a large language model with 
programmatic evaluation and iterative search. 
FunSearch~\citep{romera2024mathematical} first demonstrated that this loop can
discover genuinely new mathematical objects, outperforming decades of hand-crafted
combinatorial search.
AlphaEvolve~\citep{novikov2025alphaevolve} generalized the principle to complete
codebases, achieving state-of-the-art results on hardware scheduling, matrix
multiplication, and open problems in math and engineering.
Following this line of work, the overall procedure can be viewed as an iterative search framework in which an LLM serves as a program mutation operator that generates candidate code modifications, a task-specific evaluator assigns fitness based on predefined scoring criteria, and a selection mechanism iteratively preserves and refines high-performing programs within the search space.

Despite these successes, three interrelated bottlenecks hinder effective evolutionary search, especially in reliably steering program mutations toward high-quality regions of a large and complex program space. \textbf{(1) Reward sparsity.}
Each fitness evaluation requires a complete program execution as a full training run in the case of program search, or a numerical integration in symbolic regression. The archive is updated only when a child outperforms the current occupant of its MAP-Elites cell, so the search process receives a single scalar signal indicating whether the candidate improves upon the incumbent in terms of the defined fitness score.
Programs that nearly beat an incumbent, for example achieving performance very close to the current best, or that contain partially correct or structurally meaningful sub-expressions, are still discarded if they do not strictly improve the cell.
As a result, a large population of evaluated programs is reduced to a sparse sequence of binary improvement events, discarding fine-grained relational information about proximity to improvement and the contribution of intermediate structural modifications. \textbf{(2) Uninformed parent selection.}
AlphaEvolve samples parents by fitness rank with random diversity, treating the population as an unstructured bag of programs. However, the program population may carry substantial structural information. Effective program evolution depends on the existence of local structural similarity in the program space.  rather than merely high-level fitness comparison. Without such locality, small mutations induce unpredictable changes in program behavior, turning the search process into a largely stochastic exploration. 
In contrast, when programs exhibit meaningful local similarity, neighboring programs tend to induce more consistent and gradual changes in functionality, allowing iterative mutations to progressively refine performance. \textbf{(3) Static mutation operator.}
In AlphaEvolve, the LLM for mutation is fixed throughout the entire search process, which constrains the space of program edits to a static and pre-trained distribution. As a result, the system continues to sample from the same set of structural transformations regardless of the accumulated search experience.

\paragraph{Our contribution.}
We present \textbf{G}raph-\textbf{A}ugmeted \textbf{E}volution,
a framework that simultaneously resolves all three limitations within the OpenEvolve backbone. A \textbf{relational GNN} parses each program into a typed computation graph and
is trained online to predict fitness~(\S\ref{sec:graph_gnn}). A \textbf{Discrete SAC meta-controller} takes each program's GNN embedding
and learns which structural edit to apply next, balancing task-score
improvement against population novelty and complexity via automatic entropy
regularization~(\S\ref{sec:bandit}). A \textbf{GRPO fine-tuning loop} uses group-normalized evaluation rewards and the
parent's GNN embedding as a variance-reduction baseline to update the LLM
mutation operator online, progressively sharpening mutation quality throughout
search~(\S\ref{sec:grpo}). We evaluate GAE on an important scientific task—symbolic regression for nonlinear oscillator systems. We utilize the benchmark datasets from LLM-SR \citep{shojaee2024llm} for our experiments. 
In LLM-SR~(\S\ref{sec:experiments}) experiment,
GAE discovers novel programs that consistently match or surpass prior baselines on the benchmarks, achieving competitive or state-of-the-art performance. The remainder of the paper is organized as follows: \S\ref{sec:related} surveys related work, \S\ref{sec:method} describes the GAE framework in full, \S\ref{sec:experiments} presents experimental results,
\S\ref{sec:conclusion} discusses limitations and future directions.
\section{Related Work}
\label{sec:related}
Prior work addresses at most one of the above-mentioned bottlenecks in isolation. 
EvoTune~\citep{surina2025algorithm} finetune the LLM via DPO and GRPO respectively, yet leave parent sampling uninformed.
Surrogate-based optimization~\citep{white2021bananas,falkner2018bohb} improves evaluation
efficiency but typically operates over fixed, hand-encoded representations and does not
support open-ended code evolution. Quality-Diversity RL methods~\citep{nilsson2021policy,faldor2023map,batra2023proximal} replace random selection with policy-gradient updates, but require dense per-step rewards
unavailable when fitness arrives only at program termination. No existing method closes all three gaps simultaneously.

GAE builds on LLM-driven evolutionary search but departs from its static operator
assumption. FunSearch~\citep{romera2024mathematical}, AlphaEvolve~\citep{novikov2025alphaevolve},
and OpenEvolve~\citep{openevolve2025} all freeze model weights throughout search, relying
on prompt diversity alone for exploration. EvoPrompting~\citep{chen2023evopromptinglanguagemodelscodelevel}
extends this idea through few-shot prompting, yet still fixes the selection policy. PACEvolve~\citep{yan2026pacevolve} addresses scaffold-level failure modes, such as context pollution, mode collapse, and weak collaboration, through hand-crafted rules, but leaves the mutation model and selection policy frozen. ThetaEvolve~\citep{wang2025thetaevolve} introduces test-time RL fine-tuning to improve mutation quality, yet ties the weight update to a single task-specific objective rather than a general reusable selector. ShinkaEvolve~\citep{lange2025shinkaevolve} improves sample 
efficiency via a bandit-based LLM ensemble and novelty-based rejection filtering, yet 
the selection policy remains static and prompt-driven with no learned component.
Our work adds a learned selector and fine-tuning loop on top of this backbone, so both
the \emph{which-to-mutate} and \emph{how-to-mutate} decisions improve over time.

Maintaining population diversity while improving selection is precisely the goal of
quality-diversity RL, yet existing methods expose a fundamental mismatch with program evolution. MAP-Elites~\citep{mouret2015illuminating} archives diverse elites but samples
parents uniformly. Policy gradient extensions PGA-MAP-Elites~\citep{nilsson2021policy},
DCG-MAP-Elites~\citep{faldor2023map}, DCRL-MAP-Elites~\citep{faldor2025synergizing},
and PPGA~\citep{batra2023proximal} improve this but all require dense per-step rewards
absent in whole-program evaluation. AURORA~\citep{grillotti2022unsupervised} relaxes
hand-specified feature axes by learning behavioural descriptors unsupervisedly. GAE
adopts this spirit for archive indexing while bridging the dense-reward gap through
graph-structured signals propagated across the population.

\section{Methodology}
\label{sec:method}


\subsection{Background: MAP-Elites and OpenEvolve}
\label{sec:background}
\vspace{-4pt}

\paragraph{MAP-Elites.}
Let $\mathcal{X}$ be a program space and $\mathcal{B}$ a $k$-dimensional behaviour
descriptor space binned into cells $\mathcal{C}$.
MAP-Elites maintains an archive $\mathcal{A}: \mathcal{C} \to \mathcal{X} \cup \{\emptyset\}$
mapping each cell to its current elite program.
The QD-score aggregates quality and coverage:
\[
  \QD(\mathcal{A}) = \sum_{c \in \mathcal{C}} f(\mathcal{A}[c]) \cdot \mathbf{1}[\mathcal{A}[c] \neq \emptyset].
\]
The deterministic update rule inserts a new program $\beta$ into $\mathcal{A}$ only if
it strictly improves its cell's incumbent, guaranteeing $\QD$ is non-decreasing.

\paragraph{OpenEvolve.} OpenEvolve maintains a program archive $\mathcal{A}$ as a quality--diversity population, using a MAP-Elites scheme in which programs are mapped onto a multi-dimensional feature grid and the best performer is retained per cell~\cite{openevolve}. At each generation, it samples a parent from
$\mathcal{A}$, assembles an LLM prompt from the parent code and top-$k$ elites (drawn separately from the parent, so inspiration programs differ from those
shown to the LLM), and queries the LLM for a mutated child via SEARCH/REPLACE diffs or full rewrites. The child is scored by a user-supplied evaluator under a cascade pattern of multi-stage validation and admitted into $\mathcal{A}$ only if it dominates its cell's occupant. For diversity at scale, $\mathcal{A}$ is partitioned into multiple islands that evolve independently with periodic migration to prevent premature convergence, all orchestrated in parallel by a central controller.


\subsection{Graph-Augmented Evolution}
\label{sec:overview}
\begin{figure*}[t]
    \centering
    \includegraphics[width=\linewidth]{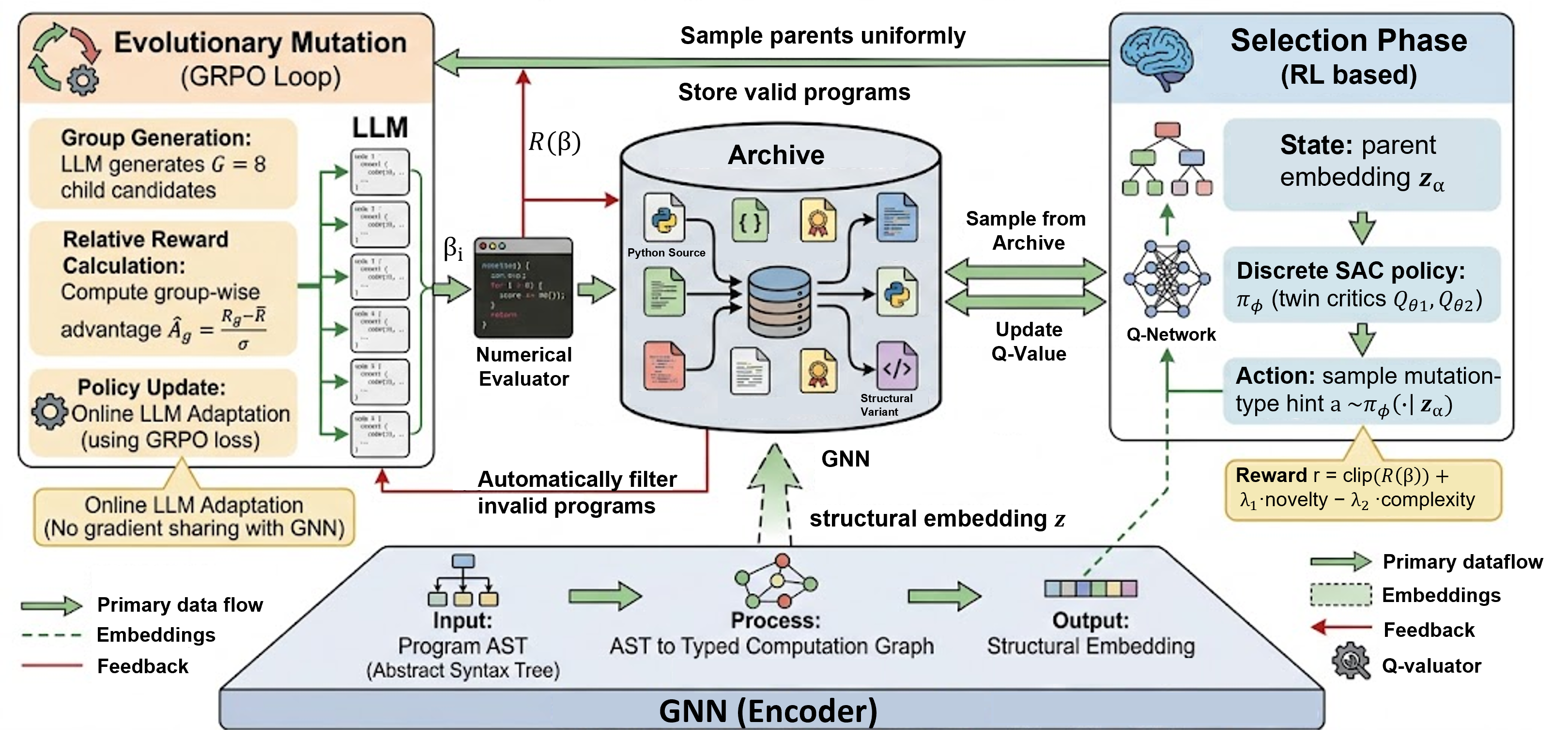}
    \caption{\textbf{GAE framework overview.} A relational GNN encoder parses each program's abstract syntax tree into a typed computation graph and produces a structural embedding $z$, fine-tuned online against observed rewards to track the evolving program distribution. Parents are sampled from a fixed-size elite archive; the parent embedding $z_\alpha$ then serves as the state of a Discrete Soft Actor--Critic meta-controller (policy $\pi_\phi$ with twin critics $Q_{\theta_1},Q_{\theta_2}$), which samples a mutation-type hint $a\sim\pi_\phi(\cdot\mid z_\alpha)$ that steers the mutation. Each parent is mutated by a local LLM generating $G=8$ candidates, which are scored by a numerical evaluator.
    The SAC policy is trained on the best child of each generation with reward $r=\mathrm{clip}(R(\beta))+\lambda_1\cdot\text{novelty}-\lambda_2\cdot\text{complexity}$, while the children's group-relative rewards adapt the LLM generation policy via GRPO.}
    \label{fig:gae_scheme}
\end{figure*}

\vspace{-4pt}
Graph-Augmented Evolution extends the
OpenEvolve loop with three co-adaptive components, all underpinned by a shared
relational GNN encoder (\S\ref{sec:graph_gnn}) that parses each program's
abstract syntax tree into a typed computation graph and produces a structural
embedding $z \in \mathbb{R}^{d}$.
Parents are drawn uniformly from a fixed-size elite archive; the shared
embedding then conditions a discrete Soft Actor--Critic meta-controller
(\S\ref{sec:bandit}) that maps the parent state $z$ to a mutation-type hint,
steering the LLM toward a structurally promising class of edit rather than an
undirected one. Its policy is trained on a three-component reward
(task score, embedding-space novelty, and a complexity penalty) computed on the
best child of each generation.
Whenever the LLM proposes a group of $G$ children, online GRPO
fine-tuning (\S\ref{sec:grpo}) converts their group-relative advantages
$\hat{A}_i \propto R_{\beta_i} - R_\alpha$ (child reward minus parent reward)
into an in-place PPO-clip update of the LLM weights, adapting the mutation
policy without interrupting evolution while leaving the archive and the GNN
unchanged.
The GNN encoder itself is updated separately every $K$ generations
via an MSE surrogate loss that regresses its embeddings onto observed rewards,
after which all archive embeddings are refreshed. The three components do not share gradients and operate at different timescales, keeping each module independently replaceable. Figure~\ref{fig:gae_scheme} illustrates the complete architecture, and Algorithm~\ref{alg:gae} summarizes one generation of the GAE loop.

\begin{algorithm}[t]
\caption{GAE --- One Generation}
\label{alg:gae}
\begin{algorithmic}[1]
\Require archive $\mathcal{P}$ (fixed size $N$); GNN encoder
  $\Phi:\text{code}\!\to\!\mathbb{R}^{d}$ ($d{=}128$); SAC policy $\pi_\phi$
  over action space $\mathcal{A}$ with twin critics $Q_{\theta_1},Q_{\theta_2}$;
  LLM operator $\pi_{\theta_{\mathrm{LLM}}}$ with reference $\pi_{\mathrm{ref}}$;
  replay buffer $\mathcal{B}$; GNN window $\mathcal{L}$;
  group size $G$; GNN interval $K$; warm-up $W$; generation index $s$.

\Statex \textit{// Parent selection and mutation hint}
\State $\alpha \sim \mathrm{Uniform}(\mathcal{P});\quad z_\alpha \gets \Phi(\alpha)$
\State $a \gets \mathrm{Uniform}(\mathcal{A})$ if $s<W$, else $a\sim\pi_\phi(\cdot\mid z_\alpha)$

\Statex \textit{// Generate and evaluate a group of $G$ children}
\State $\{\beta_1,\dots,\beta_G\}\sim \pi_{\theta_{\mathrm{LLM}}}(\cdot\mid \alpha, a)$
\State $R_{\beta_i}\gets \textsc{Eval}(\beta_i)=-\log_{10}\mathrm{NMSE}(\beta_i),\;\; i=1,\dots,G$

\Statex \textit{// GRPO update of the LLM operator (\S3.5)}
\State $\hat A_i \gets \big(R^{\mathrm{grp}}_i-\overline{R^{\mathrm{grp}}}\big)/(\mathrm{std}(R^{\mathrm{grp}})+\varepsilon)$,
  \; $R^{\mathrm{grp}}_i=R_{\beta_i}-R_\alpha$
\State Update $\theta_{\mathrm{LLM}}$ via the PPO-clip surrogate $\mathcal{L}_{\mathrm{GRPO}}$ (Eq.~8)

\Statex \textit{// SAC meta-controller update, best child only (\S3.4)}
\State $\beta^\star\gets\arg\max_i R_{\beta_i};\quad z'\gets\Phi(\beta^\star)$
\State $r \gets \mathrm{clip}(R_{\beta^\star},r_{\min},r_{\max})+\lambda_1 r_{\mathrm{nov}}(z',\mathcal{P})-\lambda_2 r_{\mathrm{cx}}(\beta^\star)$ (Eq.~6)
\State $\mathcal{B}.\mathrm{push}(z_\alpha,a,r,z')$; update $\pi_\phi,Q_{\theta_1},Q_{\theta_2}$ (Eq.~7) if $|\mathcal{B}|\ge 32$

\Statex \textit{// Archive and GNN-buffer update (all children)}
\For{$i=1,\dots,G$}
  \State $\mathcal{P}.\mathrm{Insert}(\beta_i,R_{\beta_i},\Phi(\beta_i))$, keep top-$N$ by score
  \State append $(\mathrm{graph}(\beta_i),R_{\beta_i})$ to $\mathcal{L}$ if $R_{\beta_i}>-10$
\EndFor

\Statex \textit{// Periodic GNN fine-tuning (\S3.3)}
\State \textbf{if} $s\bmod K=0$ \textbf{and} $|\mathcal{L}|\ge 10$: fine-tune $\Phi$ (Eq.~5); refresh $z_p\;\forall p\in\mathcal{P}$
\end{algorithmic}
\end{algorithm}

\subsection{Program Graphs and GNN Encoder}
\label{sec:graph_gnn}
\vspace{-4pt}

\paragraph{Graph representation.}
We parse each candidate program into a typed computation graph
$\mathcal{G} = (\mathcal{V}, \mathcal{E})$ using static Abstract Syntax Tree (AST) analysis.
Any parse failure produces a single \texttt{unknown} node, guaranteeing a valid
graph is always returned.
Specifically for symbolic regression, nodes are
mathematical operations and edges encode data-flow.

\paragraph{Relational GNN encoder.}
Each node carries an $F$-dimensional one-hot encoding its node type. We embed each program graph into $z \in \mathbb{R}^{d}$  via a
relational graph neural network. Suppose we denote the archive size as $N$, the \textbf{input projection is}
\begin{equation}
  h^{(0)} = \mathrm{GELU}(X W_{\mathrm{in}} + b_{\mathrm{in}}),
  \quad X \in \mathbb{R}^{N \times F},\;
  W_{\mathrm{in}} \in \mathbb{R}^{F \times h}.
  \label{eq:input_proj}
\end{equation}
with hidden width $h$ (the output dimension is produced only by the final pooling projection, where $2h = d$).

\paragraph{Relational message passing.}
We apply $L$ layers of edge-type-conditioned convolution. For each relation $r \in \mathcal{E}$, a dedicated weight matrix
$W_r \in \mathbb{R}^{h \times h}$ transforms the features of the source node. Messages are mean-aggregated at each destination and combined with a self-loop:
\begin{multline}
  h_i^{(\ell+1)} = \mathrm{LayerNorm}\!\left(\mathrm{GELU}\!\left(
    W_{\mathrm{self}}\, h_i^{(\ell)}\right.\right. \\
    \left.\left.+\; \sum_{r=1}^{|\mathcal{E}|}
      \frac{1}{|\mathcal{N}_r(i)|}
      \sum_{j \in \mathcal{N}_r(i)} W_r\, h_j^{(\ell)}
  \right)\right),
  \label{eq:relconv}
\end{multline}

where $\mathcal{N}_r(i)=\{\,j : (j,i)\in\mathcal{E}_r\,\}$ denotes the set of
neighbors of node $i$ connected by an edge of relation type $r$, and
$|\mathcal{N}_r(i)|$ is its cardinality.

\paragraph{Global pooling.} Mean and max pooling are concatenated and projected to the output representation $z$ as:
\begin{align}
     z &= \mathrm{GELU}\!\left(\mathrm{LayerNorm}\!\left(
    [\bar{h};\,\hat{h}]\, W_{\mathrm{out}}
  \right)\right),\\
  \bar{h} &= \tfrac{1}{N}\textstyle\sum_i h_i^{(L)},\;
  \hat{h} = \max_i h_i^{(L)}.
  \label{eq:pool}
\end{align}

\paragraph{Pretraining and online fine-tuning.}
Before evolution begins, the encoder is pretrained on synthetic programs with structural proxy scores (feature coverage, operator diversity, complexity), using a linear score-prediction head trained jointly with the encoder. During evolution, every 3 steps (once $\geq 10$ samples exist), the encoder is fine-tuned for 300 gradient steps on a sliding window of the 300 most recent $(z_\alpha, R(\alpha))$ pairs,  where $z_\alpha \in \mathbb{R}^d$ is the GNN embedding of program $\alpha$ and $R(\alpha) \in \mathbb{R}$ is its evaluator score. Targets are standardized to zero mean and unit variance. All population embeddings are then refreshed. 
The training objective is: 
\begin{equation}
  \mathcal{L}_{\mathrm{GNN}} =
  \frac{1}{|\mathcal{L}|}\sum_{\alpha \in \mathcal{L}}\!\left(\mu_\alpha - R(\alpha)\right)^2,
  \label{eq:gnn_loss}
\end{equation}
where $\mu_\alpha = \mathrm{MLP}(z_\alpha)$ is a scalar score prediction from a
temporary head trained jointly with the encoder.

\subsection{RL-Guided Mutation Selection}
\label{sec:bandit}
\vspace{-4pt}
At each evolution step a parent $\alpha$ is drawn \emph{uniformly at random}
from the current island; the reinforcement-learning agent does not select
parents. Instead, a Discrete Soft Actor-Critic (SAC) agent chooses which
\emph{mutation} to apply to $\alpha$, conditioned on the parent's graph
embedding.

\paragraph{State and action space.}
The state is the GNN embedding $z_\alpha \in \mathbb{R}^{d}$ of the
parent program. The action space is flat and fixed, with
$|\mathcal{A}|$ discrete actions, each encoding a
tuple (mutation type, target node id, argument index).

\paragraph{Networks.}
The policy $\pi_\phi(a \mid z)$ and the twin critics
$Q_{\theta_1}, Q_{\theta_2}: \mathbb{R}^{d} \to \mathbb{R}^{|\mathcal{A}|}$ are
three-layer MLPs with LayerNorm and GELU activations that score every action
from the state:
\[
  z \xrightarrow{\;256\;} \mathrm{LayerNorm} \to \mathrm{GELU}
    \xrightarrow{\;256\;} \mathrm{GELU} \to \mathbb{R}^{|\mathcal{A}|}.
\]
Each critic is paired with a target network updated by Polyak averaging
(coefficient $\tau$).

\paragraph{Action selection.}
During a warm-up of $W$ steps, actions are sampled uniformly at random to
seed the replay buffer with diverse transitions. Thereafter, the agent samples an
action stochastically from the softmax policy $\pi_\phi(a\mid z_\alpha)$ and no
policy update is performed until the buffer holds at least $32$ transitions.

\paragraph{Reward.}
The mutation producing the best child $\beta$ receives a three-component reward:
\begin{align}
  r &= \underbrace{\mathrm{clip}\bigl(R(\beta),\,r_{\min},\,r_{\max}\bigr)}_{r_{\text{task}}}
     + \underbrace{\lambda_1\bigl(1 - \min_{p\in\mathcal{P}}\cos(z_\beta, z_p)\bigr)}_{r_{\text{novelty}}}
     \nonumber \\
    &\quad - \underbrace{\lambda_2\,\max\!\Bigl(0,\,\tfrac{N(\beta)-20}{20}\Bigr)}_{r_{\text{complexity}}},
  \label{eq:reward_sac}
\end{align}
where $R(\beta)$ is the evaluator score of the best child, $\mathcal{P}$ is the
current population, $N(\beta)$ is the AST node count of $\beta$'s return
expression. The task term uses the child's absolute (clipped) score
rather than a parent-relative improvement.

\paragraph{Online SAC training.}
Each transition $(z_\alpha, a, r, z_\beta)$ is stored in a prioritized replay
buffer of capacity. Every generation we take $K$ gradient steps
(equal to the number of children per generation) on priority-sampled
mini-batches, updating the twin critics, the policy, and the entropy
temperature $\alpha$ (automatically tuned toward a target entropy of
$-\tfrac{1}{2}\log|\mathcal{A}|$). The critics minimize an importance-weighted
TD loss
\begin{align*}
  \mathcal{L}(\theta_i) &=
    \mathbb{E}_{(z,a,r,z')\sim\mathcal{B}}\!\left[
      w \,\bigl(Q_{\theta_i}(z,a) - y\bigr)^2
    \right], \\
  y &= r + \gamma\, V(z'),
  \label{eq:sac_loss}
\end{align*}
with bootstrapped value
\begin{equation}
    V(z') = \sum_a \pi_\phi(a\mid z')\bigl(\min_i Q_{\bar\theta_i}(z',a)
- \alpha\log\pi_\phi(a\mid z')\bigr)
\end{equation}
and importance weight $w$. The policy maximizes the standard entropy-regularized
objective
\begin{align*}
    J(\pi) = \mathbb{E}_{z\sim\mathcal{B}}\!\biggl[
    \sum_a \pi_\phi(a\mid z)\bigl(&\min_i Q_{\theta_i}(z,a) \\
    &- \alpha\log\pi_\phi(a\mid z)\bigr)
  \biggr].
\end{align*}

\subsection{GRPO LLM Fine-Tuning}
\label{sec:grpo}
\vspace{-4pt}
On every evolution, we construct a group of $G$ children from the sampled parent $\alpha$, $\beta_g \sim \piLLM(\cdot\mid\alpha, a), \ g=1,\ldots,G$, with $a$ denoting the action selected by the SAC meta-controllor. A GRPO update is performed immediately after the group is evaluated. Every valid child is truly evaluated, yielding rewards ${R_g}$ and malformed completions are discarded before evaluation.

The group reward is baseline-corrected by the parent's own (true) evaluated
score $R_{\alpha}$: $R^{\mathrm{grp}}_{g} = R_{g} - R_{\alpha}$, i.e.\ each
child's reward is measured as its improvement over the parent. If $\mathrm{std}(R^{\mathrm{grp}}) < 10^{-8}$ (all rewards identical), the update
is skipped. The GRPO objective fine-tunes $\theta_{\mathrm{LLM}}$ with a PPO-clip surrogate
and KL penalty $\beta_{\mathrm{kl}} = 0.1$ to the reference model:
\begin{equation}
\begin{aligned}
    \mathcal{L}_{\mathrm{GRPO}}(\theta_{\mathrm{LLM}})
    = {}& -\frac{1}{G}\sum_{g}
        \min\!\bigl(\rho_g \hat A_g,\;
        \mathrm{clip}(\rho_g, 1\pm\epsilon_{\mathrm{clip}})\,\hat A_g\bigr) \\
      & + \beta_{\mathrm{kl}}\,\bigl[D_{\mathrm{KL}}(\piLLM \,\|\, \piLLM^{\mathrm{ref}})\bigr],
\end{aligned}
\label{eq:grpo}
\end{equation}
where  $\rho_g = \exp \,\!\bigl(\log\piLLM(\beta_g|\alpha) -
\log\piLLM^{\mathrm{ref}}(\beta_g|\alpha)\bigr)$, $\epsilon_{\mathrm{clip}}$ is PPO clipping range, and the group-normalized advantage of child $g$
\begin{equation}
  \hat A_g = \frac{R^{\mathrm{grp}}_g - \bar{R}^{\mathrm{grp}}}
                  {\mathrm{std}(R^{\mathrm{grp}}) + \epsilon}.
  \label{eq:grpo_adv}
\end{equation}
The KL term is therefore evaluated for monitoring only and contributes no
gradient. The update of $\theta_{\mathrm{LLM}}$ comes solely from the PPO-clip
surrogate, with clipping to the trust region
$[1-\epsilon_{\mathrm{clip}},\,1+\epsilon_{\mathrm{clip}}]$ providing the
effective regularization and $\beta_{\mathrm{kl}}$ reported for diagnostics.

Completions with reward below $-10$ (evaluator crash sentinels) are dropped before
computing advantages. After each group, all evaluated children are inserted into the fixed-size elite population (a top-$K$ list sorted by score, with the best child inserted first for elitism), and the population is truncated to its maximum size.

\section{Experiments}
\label{sec:experiments}

To evaluate GAEvolve's capability of scientific discovery, we test its performance
on the Nonlinear Oscillators task from LLM-SR.
Nonlinear damped oscillators are described by differential equations that capture
the complex interaction among an oscillator's position, velocity, and the acting
forces.
The goal is to discover a closed-form expression $\hat{f}(x, t, v; \theta)$ that predicts the acceleration $\ddot{x} = dv/dt$ of a nonlinear harmonic oscillator, where $x$ is position, $t$ is time, $v$ is velocity.
We evaluate GAEvolve on discovering the governing equation of a synthetically
generated variant by minimizing the Normalized Mean Squared Error (NMSE) between
predicted and observed trajectories.
The actual evolved program expresses a candidate symbolic equation as executable code.
The initial program is a naive linear model
$\hat{\dot{v}} = p_0 x + p_1 t + p_2 v + p_3$, parsed into an symbolic regression expression tree via the graph builder, producing a 4-node graph with \texttt{add}, \texttt{mul},
\texttt{input\_var} and \texttt{param} nodes.


\paragraph{Dataset.} We use the \texttt{phys\_osc} (nonlinear harmonic oscillator) domain of the LSR-Synth subset of LLM-SRBench~\cite{shojaee2025llm}, predicting the acceleration
$\dot{v}$ from $(x, t, v)$. Each task supplies three disjoint sets: a training
set, an in-distribution (ID) test set, and an out-of-distribution (OOD) test set
sampled from a different input time region. For every candidate equation, the numeric
constants are optimized by BFGS to minimize the mean squared error (MSE) on the training set. The candidate is then scored by $-\log_{10}(\mathrm{NMSE})$
on the held-out ID test set, which serves as the evolutionary fitness. 
We report the median NMSE on the ID
and OOD test sets, and the latter quantifying extrapolation.


\paragraph{Experimental Setup.} GAE maintains a fixed-size elite population of 50 programs sorted by fitness. Parents are sampled uniformly from this population, and a discrete Soft Actor-Critic meta-controller (with 32 random warm-up steps) selects the mutation-type hint passed to the LLM. Each generation samples eight LLM-generated candidates in parallel. Before committing any LLM-generated candidate to the expensive numerical optimization pipeline, we apply a lightweight four-stage validation chain that filters out malformed programs early. We (i) extract valid Python code from the raw LLM output, (ii) verify syntactic correctness, (iii) sandbox the candidate by stripping unsafe statements, and (iv) execute a fast smoke test on randomly generated dummy inputs to catch runtime errors. Candidates failing any stage are discarded immediately, preventing invalid programs from entering the evaluator or corrupting the reinforcement learning replay buffer.

For the OpenEvolve baseline, the MAP-Elites archive is two-dimensional (expression tree depth, number of distinct AST node types as structural diversity), with a population of 200 programs across 4 isolated islands and periodic migration every 15 generations. Parent sampling uses fixed elite/exploration/exploitation, each at the ratio of 0.3/0.4/0.3. 
For the PACEvolve baseline, We use the official implementation with a single island. The program database maintains a score-sorted queue of at most 100 programs, and each parent is drawn by tournament selection of size 2 from the top-4 programs. Each iteration generates one candidate with the frozen LLM backbone with temperature at 0.7 and top-p at 0.95). Before code generation, a scratch-pad step proposes a hypothesis, a repository capped at 5 ideas with periodic merging, and backtracking which restores an earlier repository state sampled from a power-law distribution. 


We use Qwen3.5-35B-A3B \cite{qwen3.5_2026} as the LLM backbone of GAE.
We fix the LLM weights for PACEvolve-Single and OpenEvolve and update it in full weights in GAE during the GRPO reinforcement learning period. The GRPO has a group size of $G=8$, and uses a KL regularization coefficient of $\beta = 0.1$ together with a clipping range of $\varepsilon = 0.2$ to stabilize policy updates. Optimization is carried out using AdamW with a learning rate of $1 \times 10^{-6}$. Fine-tuning is performed in an online manner, with updates triggered at every generation after candidate evaluation.

\begin{table}[t]
\centering
\small
\begin{minipage}{0.48\textwidth}
\centering
\begin{tabular}{lcc}
\toprule
\textbf{Method} & \textbf{Mean NMSE} & \textbf{Best NMSE} \\
\midrule
\multicolumn{2}{l}{\textit{Non-LLM baseline}} \\
uDSR~\citep{landajuela2022unified}  & $-3.95$ & $-4.06$\\
\midrule
\multicolumn{2}{l}{\textit{LLM-guided evolution}} \\
LLM-SR            & $-4.06$ &  $-4.80$\\
OpenEvolve       & $-5.40$  & $-7.11$\\
CodeEvolve        & $-4.97$ & $-7.26$ \\ 
ShinkaEvolve      & $-5.35$ & $-6.35$\\
PACEvolve-Sigle  & $-5.87$ & $-8.23$ \\
PACEvolve-Multi   & $-6.11$ & $-8.24$\\
\bottomrule
\end{tabular}
\vspace{5pt}
\caption{Benchmark results on the LLM-SR Nonlinear Oscillators task~\citep{yan2026pacevolve}, reported as $\log_{10}$ NMSE ($\downarrow$). All LLM-guided baselines run for 1000 iterations over 10 independent runs. Both the mean and the best $\log_{10}$ NMSE across runs are reported.}
\label{tab:pac-results}
\end{minipage}
\hfill
\begin{minipage}{0.48\textwidth}
\centering
\begin{tabular}{lcc}
\toprule
\textbf{Method} & \textbf{ID NMSE} $\downarrow$ & \textbf{OOD NMSE} $\downarrow$ \\
\midrule
PACEvolve-Single (Gemini) & $-5.80$ & $-6.10$ \\
OpenEvolve (Gemini)       & $-3.14$ & $-3.26$ \\
OpenEvolve (Qwen)         & $-3.88$ & $-3.94$ \\
\rowcolor{gray!10}
\textbf{GAE (ours)}       & 
\makecell{$\mathbf{-6.87}$ (Mean) \\ $\mathbf{-7.24}$ (Best)}  &
\makecell{$\mathbf{-7.10}$ (Mean) \\ $\mathbf{-7.45}$ (Best)} \\
\bottomrule
\end{tabular}
\vspace{5pt}
\caption{Comparison among PACEvolve, OpenEvolve (both 1 run) and our method GAE (3 runs; mean and best reported) on the Nonlinear Oscillators task under our evaluation setting (150 iterations), reported as $\log_{10}$ NMSE ($\downarrow$) on in-distribution (ID) and out-of-distribution (OOD) test sets.}
\label{tab:ours}
\end{minipage}
\end{table}

\paragraph{Main Results.}
  First, we present results from PACEvolve~\citep{yan2026pacevolve} under their standard high-budget evaluation protocol (1000 iterations, 10 runs), which compares against uDSR~\citep{landajuela2022unified}, LLM-SR,
  AlphaEvolve,
ShinkaEvolve~\citep{lange2025shinkaevolve}. Table~\ref{tab:pac-results} reflects the performance of existing methods under their originally reported compute regime and provides a reference for overall competitiveness in the Non-linear harmonic oscillator task. Second, we conduct a controlled, low-budget comparison under a fixed backbone setting, using 
either Qwen or Gemini 2.5 Flash as the shared base model. 
In this regime, we compare the \texttt{OpenEvolve} baseline, the strong prior method \texttt{PaceEvolve}, and our proposed method \textbf{GAE}, all run for 150 iterations to ensure identical computational budget. As shown in Table~\ref{tab:ours}, \textbf{GAE} achieves \textit{best} performance compared to both baselines, with its ID NMSE at -7.24 and OOD NMSE at -7.45 across all methods. 
\begin{figure}
    \centering
    \includegraphics[width=\linewidth]{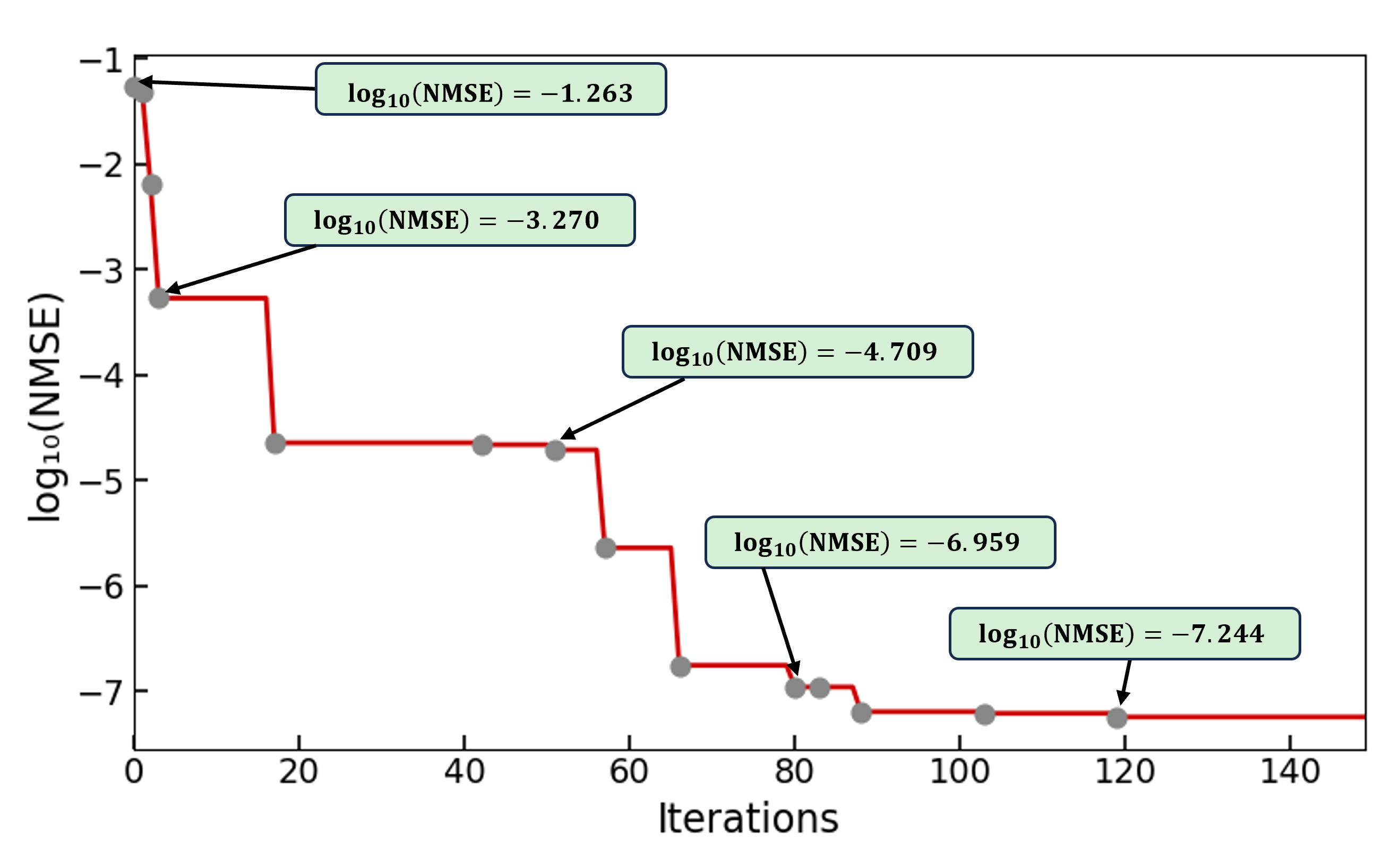}
    \caption{ Performance trajectory of the GAE method
        on the Non-linear Harmonic Oscillator symbolic regression problem.
        The curve shows the running-best $\log_{10}(\text{NMSE})$ as a function
        of the number of evolution iterations. Grey dots mark iterations at which
        a new best solution was discovered.
        Lower values of $\log_{10}(\text{NMSE})$ indicate better fit.}
    \label{fig:gae_trajectory}
\end{figure}
Figure~\ref{fig:gae_trajectory} shows the evolution trajectory of GAE over 150
iteration.
The method starts from an initial solution with
$\log_{10}(\text{NMSE}) = {-1.26}$ and improves rapidly during the first
20 iterations, reaching $\log_{10}(\text{NMSE}) \approx {-4.64}$ by iteration~17.
Steady further gains accumulate through iterations 57--88, where three
consecutive breakthroughs push the score to $\log_{10}(\text{NMSE}) = {-7.19}$.
A final improvement at iteration~119 yields the best solution of the run,
with $\log_{10}(\text{NMSE}) = {-7.24}$ (NMSE $\approx 5.7 \times 10^{-8}$),
after which the trajectory plateaus for the remaining 30 iterations.
Across all 150 iterations, 13 distinct improvements were recorded,
demonstrating that GAE efficiently discovers a near-exact symbolic form of the
target equation through graph-structured evolutionary search.

Next, we demonstrate the best program discovered by each of all four pipelines (Table~\ref{tab:ours}) and provide physical interpretations of each component in the resulting equation. \paragraph{GAE best program} 
\begin{equation}
\begin{aligned}
f(x, t, v;\,\boldsymbol{\theta}) =\;
&\theta_0\, x
+ \theta_1\, x^2\operatorname{sgn}(x)
+ \theta_2\, x^3
+ \theta_3\, x^4\operatorname{sgn}(x)\\
+\;&\theta_4\, v
+ \theta_5\, v^2\operatorname{sgn}(v)
+ \theta_6\, v^3\\
+\;&\theta_7\, \sin(t)
+ \theta_8\, \cos(t)\\
+\;&\theta_9\,\bigl(|x||v| + x^2|v| + |t||v| + t^2 v + x v^2\bigr)
\end{aligned}
\end{equation}

The terms $\theta_0 x$, $\theta_1 x^2\operatorname{sgn}(x)$, $\theta_2 x^3$, and $\theta_3 
x^4\operatorname{sgn}(x)$ correspond to the linear restoring force, quadratic and cubic 
nonlinearity, and quartic nonlinearity respectively. The terms $\theta_4 v$, $\theta_5 
v^2\operatorname{sgn}(v)$, and $\theta_6 v^3$ represent velocity-dependent damping. 
External time-dependent forcing is captured by $\theta_7\sin(t)$ and $\theta_8\cos(t)$. 
The five remaining terms $|x||v|$, $x^2|v|$, $|t||v|$, $t^2 v$, and $xv^2$, all scaled 
by the shared parameter $\theta_9$, are interaction terms. 
\paragraph{OpenEvolve best program (Qwen)} 
\[
\begin{aligned}
f(x,t,v)
=
&\;\theta_1 x
+\theta_2 x^3
+\theta_3 x^5
+\theta_4 v
+\theta_5 v^3
\\
&+\theta_6 xv
+\theta_7 x^2 v
+\theta_8 x v^2
\\
&+\theta_9 \sin t
+\theta_{10} \sin(2t)
+\theta_{11} \cos t
\\
&+\theta_{12} x\sin t
+\theta_{13} v\sin t
\\
&+\theta_{14} t
+\theta_{15} t^2
+\theta_{16} t^3
\\
&+\theta_{17}\frac{1}{1+x^2}
+\theta_{18}\frac{1}{1+v^2}
\\
&+\theta_{19} e^{-x^2}
+\theta_{20} e^{-v^2}
\\
&+\theta_{21} \sinh(x)
+\theta_{22} \cosh(x)
\\
&+\theta_{23} x^2 t
+\theta_{24} v^2 t
+\theta_{25} xvt.
\end{aligned}
\]

The first three terms ($\theta_0 x$, $\theta_1 x^3$, $\theta_2 x^5$) represent a linear, 
cubic (Duffing-like), and quintic restoring force. The next two ($\theta_3 v$, $\theta_4 v^3$) 
capture linear and cubic damping. The mixed terms $\theta_5 xv$, $\theta_6 x^2 v$, and 
$\theta_7 xv^2$ are mixed linear damping with position, mixed cubic damping ($x^2 v$), and 
mixed cubic damping ($x v^2$). Harmonic, higher harmonic, and cosine driving are handled by 
$\theta_8 \sin(t)$, $\theta_9 \sin(2t)$, and $\theta_{10} \cos(t)$, while $\theta_{11} x\sin(t)$ 
and $\theta_{12} v\sin(t)$ introduce combined time-position and time-velocity interactions. 
The terms $\theta_{13} t$, $\theta_{14} t^2$, $\theta_{15} t^3$ are polynomial time terms. 
The terms $\theta_{16}/(1+x^2)$ and $\theta_{17}/(1+v^2)$ are inverse terms for potential 
singularities. The terms $\theta_{18} e^{-x^2}$ and $\theta_{19} e^{-v^2}$ model exponential 
decay or growth in position and velocity, and $\theta_{20}\sinh(x)$ and $\theta_{21}\cosh(x)$ 
provide smooth hyperbolic transitions. Finally, $\theta_{22} x^2 t$, $\theta_{23} v^2 t$, 
and $\theta_{24} xvt$ are cross-terms.

\paragraph{OpenEvolve best program (Gemini)} 
\begin{equation}
\begin{aligned}
f(x, t, v;\,\boldsymbol{\theta}) =\;
&\theta_0\, x
+ \theta_1\, \sin(t)
+ \theta_2\, v
+ \theta_3\, x^2 v\\
+\;&\theta_4\, x^2
+ \theta_5\, v|v|
+ \theta_6\, x v^2
+ \theta_7\, x^3 \\
+ \;&\theta_8\, \cos(t)
+ \theta_9\, v^3
\end{aligned}
\end{equation}

The linear term $\theta_0 x$ represents a linear restoring force, while $\theta_1 \sin(t)$ and 
$\theta_8 \cos(t)$ together capture periodic driving or intrinsic oscillation via sine and cosine 
components. The term $\theta_2 v$ encodes linear damping. The term $\theta_3 x^2 v$ is a 
higher-order interaction combining velocity-dependent nonlinear stiffness with position-dependent 
nonlinear damping. The quadratic term $\theta_4 x^2$ introduces an asymmetric potential or 
nonlinear restoring force. Rather than $v^2$, the term $\theta_5 v|v|$ is used to model quadratic 
drag, where the damping magnitude scales with $v^2$ but the sign always opposes the direction of 
motion---a physically accurate form of turbulent or aerodynamic damping. The term $\theta_6 x v^2$ 
is a higher-order nonlinear interaction between position and velocity. The cubic term $\theta_7 x^3$ 
introduces a Duffing-oscillator-like strong nonlinear restoring force, and $\theta_9 v^3$ adds 
cubic nonlinear damping.

\paragraph{PACEvolve Best Program}

\begin{equation}
\begin{aligned}
f(x, t, v;\,\boldsymbol{\theta}) =\;
&\theta_0\, x
+ \theta_1\, v^3
+ \theta_2\, v
+ \theta_3\, x\, e^{-v^2}
+ \theta_4\, x^3\\
+\;&\theta_5\, v\, e^{-v^2}
+ \theta_6\, x v^2
+ \theta_7\, \sin(t) \\
+\;&\theta_8\, \cos(t)
+ \theta_9\, x|x|
\end{aligned}
\end{equation}

The linear term $\theta_0 x$ represents a linear restoring force, $\theta_2 v$ encodes linear 
damping, and $\theta_1 v^3$ introduces cubic nonlinear damping. The cross-term 
$\theta_3 x\, e^{-v^2}$ is a nonlinear coupling between position and velocity, where the 
Gaussian envelope $e^{-v^2}$ localizes the interaction to low-velocity regimes. The term 
$\theta_4 x^3$ adds a nonlinear restoring force of the Duffing-oscillator type. The term 
$\theta_5 v\, e^{-v^2}$ is a nonlinear damping contribution similarly localized by the Gaussian 
factor, attenuating its effect at large velocities. The higher-order cross-term $\theta_6 x v^2$ 
captures nonlinear coupling between position and the square of velocity. The terms 
$\theta_7 \sin(t)$ and $\theta_8 \cos(t)$ represent external periodic driving forces via sine 
and cosine components. Finally, $\theta_9 x|x|$ provides an additional nonlinear restoring 
force whose magnitude scales with $x^2$ but whose sign tracks the direction of displacement, 
analogous to quadratic drag in the position domain.

\section{Conclusion}
\label{sec:conclusion}

We presented \textbf{GAE}, a framework that augments LLM-guided evolutionary search with three tightly coupled components: a relational GNN encoder
that maps typed program graphs into structural embeddings, 
a contextual bandit selector that learns which program structures yield productive
mutations, and a GRPO fine-tuning loop that adapts the mutation LLM online.
Together, these components close three bottlenecks: reward sparsity, uninformed parent selection, and a static mutation
operator. Experiments on symbolic regression show that GAE
achieves the lowest NMSE both in-distribution and out-of-distribution among all baselines,
attaining state-of-the-art performance.

\textbf{Limitations.}
Our method relies on a graph representation derived from Python ASTs, which encodes domain-specific inductive biases. Extending to other languages or domains requires redesigning node types and relations, limiting out-of-the-box generalization and introducing non-trivial adaptation overhead. The benefit of graph-guided policy learning depends on a sufficiently expensive evaluation regime, as the GNN and reinforcement optimization introduce additional overhead. 

\section*{Impact Statement}



This work provides a fundamental methodological advancement in LLM-driven evolutionary algorithms. By introducing a dynamic, structure-aware paradigm—driven by relational graph encoders and online reinforcement optimization—our framework transforms evolutionary search from a largely stochastic process into a directed, self-improving trajectory. This algorithmic evolution is vital for scaling AI to tackle highly complex scientific problems, where the search spaces for viable mathematical or programmatic hypotheses are exceedingly vast and their evaluations computationally expensive.

The immediate impact of this framework is demonstrated in its ability to automate the discovery of interpretable, closed-form mathematical models from complex observational data. 
This methodology lays the technical groundwork for more comprehensive scientific agent systems designed to solve multifaceted scientific problems across theoretical physics, fluid dynamics, and complex systems engineering.

We foresee no significant harmful social consequences specific to this work. The broader social consequences of our work are those that are
well established when advancing the fields of Machine Learning and AI for
Science, and we do not feel any must be specifically highlighted here.


\bibliographystyle{abbrvnat}
\bibliography{references}

@article{shojaee2025llm,
  title={Llm-srbench: A new benchmark for scientific equation discovery with large language models},
  author={Shojaee, Parshin and Nguyen, Ngoc-Hieu and Meidani, Kazem and Farimani, Amir Barati and Doan, Khoa D and Reddy, Chandan K},
  journal={arXiv preprint arXiv:2504.10415},
  year={2025}
}

@software{openevolve,
  title = {OpenEvolve: an open-source evolutionary coding agent},
  author = {Asankhaya Sharma},
  year = {2025},
  publisher = {GitHub},
  url = {https://github.com/algorithmicsuperintelligence/openevolve}
}

@article{mouret2015illuminating,
  title={Illuminating search spaces by mapping elites},
  author={Mouret, Jean-Baptiste and Clune, Jeff},
  journal={arXiv preprint arXiv:1504.04909},
  year={2015}
}

@article{grillotti2022unsupervised,
  title={Unsupervised behavior discovery with quality-diversity optimization},
  author={Grillotti, Luca and Cully, Antoine},
  journal={IEEE Transactions on Evolutionary Computation},
  volume={26},
  number={6},
  pages={1539--1552},
  year={2022},
  publisher={IEEE}
}

@article{faldor2025synergizing,
  title={Synergizing quality-diversity with descriptor-conditioned reinforcement learning},
  author={Faldor, Maxence and Chalumeau, F{\'e}lix and Flageat, Manon and Cully, Antoine},
  journal={ACM Transactions on Evolutionary Learning},
  volume={5},
  number={1},
  pages={1--35},
  year={2025},
  publisher={ACM New York, NY}
}

@article{zhang2019d,
  title={D-vae: A variational autoencoder for directed acyclic graphs},
  author={Zhang, Muhan and Jiang, Shali and Cui, Zhicheng and Garnett, Roman and Chen, Yixin},
  journal={Advances in neural information processing systems},
  volume={32},
  year={2019}
}

@article{shahriari2015taking,
  title={Taking the human out of the loop: A review of Bayesian optimization},
  author={Shahriari, Bobak and Swersky, Kevin and Wang, Ziyu and Adams, Ryan P and De Freitas, Nando},
  journal={Proceedings of the IEEE},
  volume={104},
  number={1},
  pages={148--175},
  year={2015},
  publisher={IEEE}
}

@inproceedings{falkner2018bohb,
  title={BOHB: Robust and efficient hyperparameter optimization at scale},
  author={Falkner, Stefan and Klein, Aaron and Hutter, Frank},
  booktitle={International conference on machine learning},
  pages={1437--1446},
  year={2018},
  organization={PMLR}
}

@inproceedings{white2021bananas,
  title={Bananas: Bayesian optimization with neural architectures for neural architecture search},
  author={White, Colin and Neiswanger, Willie and Savani, Yash},
  booktitle={Proceedings of the AAAI conference on artificial intelligence},
  volume={35},
  number={12},
  pages={10293--10301},
  year={2021}
}

@article{wang2025thetaevolve,
  title={Thetaevolve: Test-time learning on open problems},
  author={Wang, Yiping and Su, Shao-Rong and Zeng, Zhiyuan and Xu, Eva and Ren, Liliang and Yang, Xinyu and Huang, Zeyi and He, Xuehai and Ma, Luyao and Peng, Baolin and others},
  journal={arXiv preprint arXiv:2511.23473},
  year={2025}
}

@article{surina2025algorithm,
  title={Algorithm discovery with llms: Evolutionary search meets reinforcement learning},
  author={Surina, Anja and Mansouri, Amin and Quaedvlieg, Lars and Seddas, Amal and Viazovska, Maryna and Abbe, Emmanuel and Gulcehre, Caglar},
  journal={arXiv preprint arXiv:2504.05108},
  year={2025}
}

@inproceedings{nilsson2021policy,
  title={Policy gradient assisted map-elites},
  author={Nilsson, Olle and Cully, Antoine},
  booktitle={Proceedings of the Genetic and Evolutionary Computation Conference},
  pages={866--875},
  year={2021}
}

@inproceedings{faldor2023map,
  title={Map-elites with descriptor-conditioned gradients and archive distillation into a single policy},
  author={Faldor, Maxence and Chalumeau, F{\'e}lix and Flageat, Manon and Cully, Antoine},
  booktitle={Proceedings of the Genetic and Evolutionary Computation Conference},
  pages={138--146},
  year={2023}
}

@article{batra2023proximal,
  title={Proximal policy gradient arborescence for quality diversity reinforcement learning},
  author={Batra, Sumeet and Tjanaka, Bryon and Fontaine, Matthew C and Petrenko, Aleksei and Nikolaidis, Stefanos and Sukhatme, Gaurav},
  journal={arXiv preprint arXiv:2305.13795},
  year={2023}
}

@article{openevolve2025,
  title   = {OpenEvolve: Open-Source Evolutionary Code Optimization with Real-World GPU Kernel Discovery},
  author  = {Sharma, Asankhaya},
  journal = {Hugging Face Blog},
  year    = {2025},
  url     = {https://huggingface.co/blog/codelion/openevolve}
}

@article{yan2026pacevolve,
  title={Pacevolve: Enabling long-horizon progress-aware consistent evolution},
  author={Yan, Minghao and Peng, Bo and Coleman, Benjamin and Chen, Ziqi and Xie, Zhouhang and Chen, Shuo and He, Zhankui and Sachdeva, Noveen and Ye, Isabella and Wang, Weili and others},
  journal={arXiv preprint arXiv:2601.10657},
  year={2026}
}

@article{shojaee2024llm,
  title={Llm-sr: Scientific equation discovery via programming with large language models},
  author={Shojaee, Parshin and Meidani, Kazem and Gupta, Shashank and Farimani, Amir Barati and Reddy, Chandan K},
  journal={arXiv preprint arXiv:2404.18400},
  year={2024}
}

@article{landajuela2022unified,
  title={A unified framework for deep symbolic regression},
  author={Landajuela, Mikel and Lee, Chak Shing and Yang, Jiachen and Glatt, Ruben and Santiago, Claudio P and Aravena, Ignacio and Mundhenk, Terrell and Mulcahy, Garrett and Petersen, Brenden K},
  journal={Advances in Neural Information Processing Systems},
  volume={35},
  pages={33985--33998},
  year={2022}
}

@article{novikov2025alphaevolve,
  title={Alphaevolve: A coding agent for scientific and algorithmic discovery},
  author={Novikov, Alexander and V{\~u}, Ng{\^a}n and Eisenberger, Marvin and Dupont, Emilien and Huang, Po-Sen and Wagner, Adam Zsolt and Shirobokov, Sergey and Kozlovskii, Borislav and Ruiz, Francisco JR and Mehrabian, Abbas and others},
  journal={arXiv preprint arXiv:2506.13131},
  year={2025}
}

@article{lange2025shinkaevolve,
  title={Shinkaevolve: Towards open-ended and sample-efficient program evolution},
  author={Lange, Robert Tjarko and Imajuku, Yuki and Cetin, Edoardo},
  journal={arXiv preprint arXiv:2509.19349},
  year={2025}
}

@misc{chen2023evopromptinglanguagemodelscodelevel,
      title={EvoPrompting: Language Models for Code-Level Neural Architecture Search}, 
      author={Angelica Chen and David M. Dohan and David R. So},
      year={2023},
      eprint={2302.14838},
      archivePrefix={arXiv},
      primaryClass={cs.NE},
      url={https://arxiv.org/abs/2302.14838}, 
}

@article{romera2024mathematical,
  title={Mathematical discoveries from program search with large language models},
  author={Romera-Paredes, Bernardino and Barekatain, Mohammadamin and Novikov, Alexander and Balog, Matej and Kumar, M Pawan and Dupont, Emilien and Ruiz, Francisco JR and Ellenberg, Jordan S and Wang, Pengming and Fawzi, Omar and others},
  journal={Nature},
  volume={625},
  number={7995},
  pages={468--475},
  year={2024},
  publisher={Nature Publishing Group UK London}
}

@misc{qwen3.5_2026,
  title={Qwen3.5 Technical Report},
  author={Qwen Team},
  year={2026},
  howpublished={\url{https://github.com/QwenLM/Qwen3.5}},
  note={Accessed: 2026-05-08}
}


\newpage
\appendix
\onecolumn
\section{GNN Architecture Details}
\label{app:gnn}

\paragraph{Input projection.}
For task $\tau$, the one-hot matrix
$X \in \{0,1\}^{N \times |\mathcal{V}_\tau|}$ is projected to hidden dimension
$h$: $h^{(0)} = \mathrm{GELU}(X W_{\mathrm{in}})$.

\paragraph{Relational convolution.}
For each relation $r \in \{1,\ldots,|\mathcal{E}|\}$, a dedicated
$W_r \in \mathbb{R}^{h \times h}$ transforms source features; messages are
mean-aggregated and summed with a self-loop $W_{\mathrm{self}}$. Each layer applies
LayerNorm and GELU (Eq.~\ref{eq:relconv}).
This is a \emph{relational GCN}, and no attention weights are computed.



\paragraph{Fallback.}
Any program failing AST parsing produces a single \texttt{unknown}/\texttt{unknown\_layer}
node; the encoder still produces a valid $z \in \mathbb{R}^{128}$.




\section{Decomposability of SAC, GRPO, and the GNN Encoder}
\label{app:decomp}

The SAC meta-controller, the GRPO fine-tuning loop, and the GNN encoder
optimize disjoint parameter sets ($\{\phi,\theta_1,\theta_2\}$,
$\theta_{\mathrm{LLM}}$, and $\theta_{\mathrm{GNN}}$ respectively). No
gradients flow between them, and each operates on its own timescale
(per generation, per group, and every $K$ generations). Disabling any
one component degrades gracefully to a simpler baseline:

\begin{center}
\small
\begin{tabular}{llll}
\toprule
\textbf{Component} & \textbf{Parameters} & \textbf{Training signal} & \textbf{Without it} \\
\midrule
SAC meta-controller & $\phi,\;\theta_1,\theta_2$ &
  $r = \mathrm{clip}(R(\beta^\star)) + \lambda_1 r_{\mathrm{nov}} - \lambda_2 r_{\mathrm{cx}}$ (Eq.~6) &
  Uniform mutation hint \\
GRPO & $\theta_{\mathrm{LLM}}$ &
  $\hat{A}_g$ from $R^{\mathrm{grp}}_g = R_g - R_\alpha$ (Eqs.~8--9) &
  Frozen LLM \\
GNN encoder & $\theta_{\mathrm{GNN}}$ &
  $\mathcal{L}_{\mathrm{GNN}}$: MSE on $\mathcal{L}$ (Eq.~5) &
  Stale embeddings \\
\bottomrule
\end{tabular}
\end{center}

Parent selection is uniform over the elite archive in all configurations
and is unaffected by any of the three components.


\section{Discrete Action Space --- Mutation Operators}
\label{app:sac-actions}

The Discrete SAC meta-controller operates over a flat, fixed action space of
$|\mathcal{A}| = 5 \times 30 \times 18 = 2700$ discrete actions, each encoding
a tuple (\emph{mutation type}, \emph{target node id}, \emph{argument index}).
The target node id indexes the parent's expression tree in BFS order (capped at
30 nodes), and the argument index selects an operator or constant from the
vocabularies below. The mutation-type component of the selected action is
additionally injected into the LLM prompt as a structural hint
(``prefer \texttt{<type>} changes''), steering generation toward the chosen
edit class. The full tuple is used when the mutation is applied directly at
the AST level (the fallback path when no valid LLM completion is produced).

\begin{table}[h]
\centering
\caption{Discrete mutation-type vocabulary for the SAC meta-controller
(symbolic regression).}
\label{tab:sac-actions}
\begin{tabular}{cll}
\toprule
\textbf{Action} & \textbf{Label} & \textbf{Description} \\
\midrule
0 & \texttt{add\_operator}     & Wrap a node with a unary op, or insert a binary op above it \\
1 & \texttt{replace\_operator} & Replace an existing operator node (arity-preserving) \\
2 & \texttt{delete\_node}      & Remove a unary wrapper, promoting its child \\
3 & \texttt{insert\_constant}  & Multiply a sub-expression by a constant \\
4 & \texttt{subtree\_rewrite}  & Replace the node's subtree with a fresh random tree \\
\bottomrule
\end{tabular}
\end{table}

\noindent
The operator vocabulary comprises 5 binary operators
(\texttt{add}, \texttt{sub}, \texttt{mul}, \texttt{div}, \texttt{pow}) and
13 unary functions
(\texttt{sin}, \texttt{cos}, \texttt{tan}, \texttt{exp}, \texttt{log},
\texttt{sqrt}, \texttt{abs}, \texttt{tanh}, \texttt{arcsin}, \texttt{arccos},
\texttt{arctan}, \texttt{sinh}, \texttt{cosh}); the constant vocabulary
contains 9 values
$\{-2, -1, -0.5, 0.5, 1, 2, 3, \pi, e\}$, giving an argument-index space of
size $\max(18, 9) = 18$. All expressions are built from the three observable
inputs: position~$x$, velocity~$v$, and time~$t$. Numeric constants
introduced as \texttt{params[$j$]} remain free parameters optimized by BFGS
during evaluation. Generated code applies numerical guards (clipped
exponents, $\epsilon$-protected division and logarithms) so that every
mutated program remains evaluable.




\section{Algorithm}

\begin{algorithm}[H]
\caption{Discrete SAC for Program Mutation}
\label{alg:sac}
\begin{algorithmic}[1]
\Require{GNN encoder $\Phi$, evaluator $\mathcal{E}$, elite population
         $\mathcal{P}$, prioritized replay buffer $\mathcal{B}$,
         warm-up $W$, total steps $S$}
\State Initialise policy $\pi_\phi$, twin critics $Q_{\theta_1}, Q_{\theta_2}$,
       targets $Q_{\bar\theta_1}, Q_{\bar\theta_2}$, temperature $\log\alpha$
\For{$s = 1, \ldots, S$}
    \State $\alpha_{\mathrm{prog}} \sim \mathrm{Uniform}(\mathcal{P})$
           \Comment{parent program, sampled uniformly}
    \State $z \leftarrow \Phi(\texttt{build\_graph}(\alpha_{\mathrm{prog}}))$
    \State $a \sim
      \begin{cases}
        \mathrm{Uniform}(\mathcal{A}) & \text{if } s < W\\
        \pi_\phi(\cdot \mid z)        & \text{otherwise}
      \end{cases}$
    \State $\beta \leftarrow \texttt{mutate}(\alpha_{\mathrm{prog}},
           \texttt{decode}(a))$
           \Comment{AST edit via \texttt{MutableSRTree}}
    \State $R(\beta) \leftarrow \mathcal{E}(\beta)$
           \Comment{BFGS constant fit $+$ NMSE}
    \State $z' \leftarrow \Phi(\texttt{build\_graph}(\beta))$
    \State $r \leftarrow \mathrm{clip} \!\!\big(R(\beta), r_{\min}, r_{\max}\big)
           + \lambda_1\, r_{\mathrm{nov}}(z', \mathcal{P})
           - \lambda_2\, r_{\mathrm{cx}}(\beta)$
           
    \State Store $(z, a, r, z')$ in $\mathcal{B}$
    \If{$|\mathcal{B}| \geq 32$ \textbf{and} $s \geq W$}
        \State Sample prioritized minibatch; update
               $Q_{\theta_1}, Q_{\theta_2}$,
               $\pi_\phi$, and $\alpha$
        \State Soft-update targets $Q_{\bar\theta_1}, Q_{\bar\theta_2}$
               with Polyak coefficient $\tau$
    \EndIf
    \State $\mathcal{P}.\mathrm{Insert}(\beta, R(\beta), z')$
           \Comment{keep top-$N$ by score}
\EndFor
\end{algorithmic}
\end{algorithm}

\section{Hyperparameters}
\label{app:hparams}

Table~\ref{tab:hparams} collects all parameters introduced in the
methodology (\S3.2--\S3.5), grouped by component. Values marked
$^\dagger$ are set in the implementation but not stated explicitly in
the main text.

\begin{table}[t]
\centering
\caption{Hyperparameters of GAE, grouped by component.}
\label{tab:hparams}
\small
\begin{tabular}{llc}
\toprule
\textbf{Parameter} & \textbf{Meaning} & \textbf{Value} \\
\midrule
\multicolumn{3}{l}{\emph{Elite archive and evolution loop (\S3.2, Alg.~1)}} \\
$N$        & Archive size (top-$N$ elite population)            & $50$ \\
$G$        & Children generated per generation (group size)     & $8$ \\
$K$        & GNN fine-tuning interval (generations)             & $3$ \\
$d$        & Structural embedding dimension                     & $128$ \\
\midrule
\multicolumn{3}{l}{\emph{GNN encoder (\S3.3)}} \\
$F$        & Node-type vocabulary size (one-hot input)          & $17$ \\
$|\mathcal{E}|$ & Number of typed edge relations                & $6$ \\
$L$        & Relational message-passing layers                  & $2$ \\
$h$        & Hidden width (pooled projection gives $2h = d$)    & $64$ \\
---        & Gradient steps per fine-tuning round               & $300$ \\
---        & Sliding-window size of $\mathcal{L}$ (recent pairs) & $300$ \\
---        & Minimum samples before fine-tuning ($|\mathcal{L}|$) & $10$ \\
\midrule
\multicolumn{3}{l}{\emph{Discrete SAC meta-controller (\S3.4)}} \\
$|\mathcal{A}|$ & Action-space size $(5 \times 30 \times 18)$   & $2700$ \\
---        & Policy/critic MLP hidden width                     & $256$ \\
$\tau$     & Polyak target-update coefficient                   & $0.005$ \\
$W$        & Warm-up steps (uniform actions)                    & $32$ \\
---        & Minimum buffer size before updates                 & $32$ \\
---        & Replay buffer capacity (prioritized)               & $10{,}000$ \\
---        & Mini-batch size                                    & $64$ \\
$\gamma$   & Discount factor in the TD target                   & $0.99^\dagger$ \\
---        & SAC learning rate (Adam, all networks)             & $3\times10^{-4}\,^\dagger$ \\
$\bar{\mathcal{H}}$ & Target entropy (auto-tuned $\alpha$)      & $-\tfrac{1}{2}\log|\mathcal{A}|$ \\
$r_{\min},\, r_{\max}$ & Task-score clipping bounds (Eq.~6)      & $-5,\; 15$ \\
$\lambda_1$ & Novelty bonus weight (Eq.~6)                      & $0.1$ \\
$\lambda_2$ & Complexity penalty weight (Eq.~6)                 & $0.05$ \\
---        & Complexity penalty onset (AST nodes, Eq.~6)        & $20$ \\
\midrule
\multicolumn{3}{l}{\emph{GRPO LLM fine-tuning (\S3.5)}} \\
$\epsilon_{\mathrm{clip}}$ & PPO clipping range (Eq.~8)          & $0.2$ \\
$\beta_{\mathrm{kl}}$ & KL coefficient (monitoring only, Eq.~8)  & $0.1$ \\
---        & Crash-sentinel reward threshold (dropped)          & $-10$ \\
---        & Advantage-std skip threshold                       & $10^{-8}$ \\
\bottomrule
\end{tabular}
\end{table}


\end{document}
